\pdfoutput=1
\documentclass[11pt]{article}

\usepackage[final]{acl}
\usepackage{longtable}

\usepackage{times}
\usepackage{latexsym}
\usepackage{hyperref}

\usepackage[T1]{fontenc}
\usepackage[utf8]{inputenc}

\usepackage{microtype}
\usepackage{float}

\usepackage{inconsolata}
\usepackage{enumitem}
\setlist{noitemsep}

\newcommand{\sysname}{{\textbf{\sc Defend}}}
\usepackage{graphicx}
\usepackage{multirow}
\usepackage{array}
\usepackage{graphicx}
\usepackage{tabularx, booktabs}
\usepackage{tikz}
\usetikzlibrary{arrows.meta, positioning, shapes.geometric}
\title{Defend: Automated Rebuttals for Peer Review with Minimal Author Guidance}

\author{Jyotsana Khatri \\
  TCS Research \\
  \texttt{khatri.jyotsana@tcs.com} \\\And
  Manasi Patwardhan \\
  TCS Research \\
  \texttt{manasi.patwardhan@tcs.com} \\%\And
  }

\begin{document}
\maketitle
\begin{abstract}
Rebuttal generation is a critical component of the peer review process for scientific papers, enabling authors to clarify misunderstandings, correct factual inaccuracies, and guide reviewers toward a more accurate evaluation.
We observe that Large Language Models (LLMs) often struggle to perform targeted refutation and maintain accurate factual grounding when used directly for rebuttal generation, highlighting the need for structured reasoning and author intervention. To address this, in the paper, we introduce \sysname\, an LLM based tool designed to explicitly execute the underlying reasoning process of automated rebuttal generation, while keeping the author-in-the-loop. As opposed to writing the rebuttals from scratch, the author needs to only drive the reasoning process with minimal intervention, leading an efficient approach with minimal effort and less cognitive load.
We compare \sysname\ against three other paradigms: (i) Direct rebuttal generation using LLM (DRG), (ii) Segment-wise rebuttal generation using LLM (SWRG), and (iii) Sequential approach (SA) of segment-wise rebuttal generation without author intervention.   
To enable fine-grained evaluation, we extend the ReviewCritique dataset, creating review segmentation, deficiency, error type annotations, rebuttal-action labels, and mapping to gold rebuttal segments. Experimental results and a user study demonstrate that directly using LLMs perform poorly in factual correctness and targeted refutation. Segment-wise generation and the automated sequential approach with author-in-the-loop, substantially improve factual correctness and strength of refutation.  
\end{abstract}

\section{Introduction}
The peer-review process is critical to ensuring the rigor, and impact of scientific research.  Reviewers provide feedback aimed at identifying factual inaccuracies, issues in methodology or experimentation, unclear writing, etc; while authors must provide a rebuttal to clarify misunderstandings, and directly address the concerns in the review, influencing the final scores, further revisions and thus future research directions  \citet{kargaran2025insights}.  
%What is a good rebuttal
An effective rebuttal must directly address the reviewer's specific concerns, factually accurate, grounded in the paper and relevant literature, maintain internal consistency, and most importantly able to refute wherever appropriate. 

Among these qualities refutation, factual correctness and consistency are the most important components, as a rebuttal that fails to correct factual inaccuracies, technical misinterpretations and unsubstantiated claims in the review can negatively influence the reviewer's decision. Authors must precisely interpret reviewer's critique, and  craft appropriate response strategies (actions) - such as clarifying, correcting, justifying, or conceding with a plan for revision, while grounding the response in the paper and the literature. Rebuttal generation is a cognitively demanding task and must be carried out within a limited strictly defined time frame. 

%Our observations and approach
Our initial experiments with zero-shot LLM prompting demonstrates the inability of proprietary LLMs to identify deficient review segments and generate rebuttals with \textbf{correct and consistent factual claims, and targeted refutation}.  It hints at the fact that 
there is a need to decompose the complex task into multi-step reasoning  process \cite{wei2022chain,wangself}  
We observe that, with an appropriate action serving as a cue, LLMs can generate appropriate rebuttals. We further observe that providing LLMs with explicit cues about the appropriate response in terms of deficiency, error-type, and rebuttal action significantly improves the quality of generated rebuttals.

We introduce \sysname\, an LLM augmented tool designed to explicitly simulate %execute 
the underlying reasoning process of automated  rebuttal generation, while keeping author-in-the-loop. As opposed to writing the rebuttals from scratch, the author needs only drive the reasoning process with minimal intervention, leading an efficient approach with less cognitive load. We compare the outputs generated by Defend with (i) Direct Rebuttal Generation \textbf{(DRG)} with zero-shot LLM prompting using the paper content, and the entire review as input,
(ii)  Segment-wise Rebuttal generation \textbf{(SWRG)} with zero-shot LLM prompting using each semantic review segment as an input following by consolidation to formulate complete rebuttal, and (iii) Sequential Approach \textbf{(SA)}, which performs complete automation of the rebuttal generation including Review Deficiency Prediction (DP), Error Type Prediction (ETP), Rebuttal Action Prediction (RAP), and final Rebuttal Generation (RG). 
To enable systematic evaluation of the tool, we  extend the ReviewCritique\cite{du2024llms} dataset to map gold rebuttal for each review segment and label it with a rebuttal action.
Experimental results show that DRG performs poorly in factual correctness and targeted refutation, whereas both SWRG and \sysname\ significantly improve factual grounding and strength of refutation. However, the stages of the SA pipeline, remain prone to mis-predictions, underscoring the need for author-in-the-loop correction. Our interactive tool \sysname\, that first presents SWRG-generated rebuttal drafts and, when necessary, takes author's guidance throughout the pipeline for iterative refinement, lead to near gold rebuttal.

The main contributions of this work are:
\begin{itemize}
    \item We create an evaluation set by extending a subset of ReviewCritique  \cite{du2024llms} including fine-grained annotations of deficiency, error type,  rebuttal action and gold rebuttal for each semantic review segment.
    \item We have built an interactive tool \sysname\ to assist authors in AI augmented rebuttal generation.
    \item We  benchmark each fine-grained step of the sequential  \sysname\ pipeline against DRG, SWRG, and SA, to demonstrate the need of to-the-point author interventions with very less cognitive load, at the same time generating good rebuttals from the perspective of capability of refutation, factual correctness and consistency.
\end{itemize}

\section{Related work}

Early efforts on automation of peer-review life-cycle were focused on creating datasets of collection of papers and their corresponding reviews \cite{kang2018dataset,dycke2023nlpeer,zhang2022investigating,gao2024reviewer2,zhou2024llm,zhu2025deepreview, Bharti2023PolitePEERDP,lin2023moprd}.
%PeerRead\cite{Kang2018ADO}, MOPRD\cite{lin2023moprd}, NLPeer\cite{dycke2023nlpeer} contains a large number of peer-reviews for papers, without any fine-grained expert annotations for reviews or rebuttals.
These datasets enabled a range of downstream tasks including score and acceptance prediction \cite{Kang2018ADO,Bharti2021PEERAssistLO,Dycke2022NLPeerAU,du2024llms}, review and meta-review generation \cite{Shen2021MReDAM,Zhang2022InvestigatingFD,Yuan2021CanWA,Wu2022IncorporatingPR,DArcy2023ARIESAC,jin2024agentreview,Zhou2024IsLA,Weng2024CycleResearcherIA,Zhu2025DeepReviewIL}, review analysis \cite{kennard2022disapere}, argument mining \cite{hua2019argument,ruggeri2023dataset,cheng2020ape}, review-rebuttal discussions \cite{wu2022incorporating,tan2024peer} and review-informed paper revision \cite{DArcy2023ARIESAC}. The ReviewCritique \citet{du2024llms} dataset provides a benchmark to identify whether a review segment is deficient (have misunderstandings, factual errors, or wrong) with fine-grained annotations about the type of errors made by the reviewer.

Subsequent work began incorporating the rebuttal data to improve the overall automated peer-review process together with fine-grained annotations \cite{kennard2022disapere,Purkayastha2023ExploringJA,jin2024agentreview,wu2022incorporating,tan2024peer}. 
\citet{Purkayastha2023ExploringJA} explores the jiu-jitsu argumentation for peer-review process by proposing attitude and theme guided rebuttal generation. However, they do not focus on the part when a review segment is deficient. The DISAPERE corpus \cite{kennard2022disapere} focused on providing discourse level annotations for review-rebuttal discussions. \citet{zhang2025re} released largest dataset of peer-review and rebuttals, they showed results with zeroshot and finetuned models to simulate review-rebuttal discussions. 
None of these works provides mapping of reviews and rebuttals at fine-grained segment level, having intermediate annotations in terms of deficiencies in the review segment and the corresponding action taken by the author to address those deficiencies leading  to the  corresponding rebuttal segment. For evaluation of \sysname\ we have come-up with a dataset with such fine-grained annotations. \cite{anonymous2025do} shows that LLMs performs well for meta-review summarization by re-organizing the existing arguments in the review, but not for rebuttal generation by the creation of new arguments, which substantiates our claim of the need for author-intervention for the rebuttal generation task.

\section{Methodology}

\subsection{Direct Rebuttal Generation  (DRG)}
An LLM is provided with paper title, paper content, and the complete review. The model is then tasked with generating a rebuttal directly in one step, without any intermediate reasoning, planning, or decomposition. %(Prompt in Table \ref{tab:prompts_conso_drg} in appendix).

\subsection{Segment-wise Rebuttal Generation (SWRG)}
We prompt LLM to segment the review based on meaningful components such as summary of the paper, strengths, weaknesses (each strength/ weakness is a separate segment) etc.  %(Prompt in Table \ref{tab:segmentation} in appendix ).
For each segment, a rebuttal is generated with an LLM call %((Table \ref{tab:prompts_swrg}) in appendix) 
containing information about the paper title, paper content, and the entire review in context. %Once a rebuttal is generated for all segments, 
We further consolidate the rebuttal segments to create a final rebuttal.% (Prompt in Table \ref{tab:prompts_conso_drg}  in appendix).

\subsection{Sequential approach (SA)}
\begin{figure*}[!t]
  \centering
  \includegraphics[width=\textwidth,scale=.8]{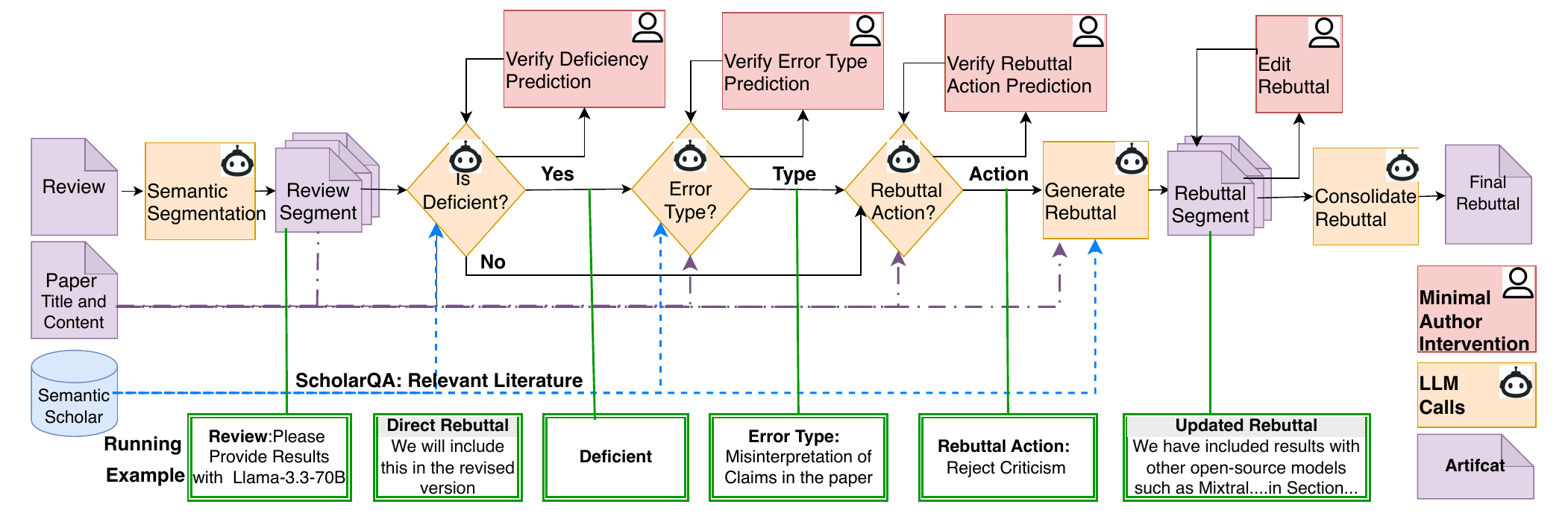}
  \caption{Defend: Rebuttal Generation Pipeline}
  \label{fig:DemoPipeline}
\end{figure*}

For writing a rebuttal an author first forms semantic units of a review. 
For segments which has issues in terms of factuality, subjectivity, tone etc., the author would mostly reject the demand or clarify the misconceptions. \cite{du2024llms} calls such segments deficient and divide them into fine-grained error types.
We explicitly simulate this step-wise reasoning process an author follows, leading to a sequential rebuttal generation pipeline which contains 4 steps (i) Deficiency prediction, (ii) Error-type prediction, (iii) Rebuttal action prediction, and (iv) Rebuttal generation. Figure \ref{fig:DemoPipeline} shows the flow of pipeline. 

\textbf{Deficiency Prediction (DP):}
An LLM is prompted with paper title, paper content, and the entire review together with the review segment which is under consideration and definition of deficiency  \cite{du2024llms}, to determine the deficiency status of a review segment, which is a binary classification task.

\textbf{Error-type Prediction (ETP):}
For review segments classified as deficient, an LLM call is made to determine the error type, which denotes a more fine-grained classification of the exact problem in the deficient review segments. We use the same error-types provided in \cite{du2024llms} but with more coarse-grained classification to group the similar error-types. 
We opted for a coarser taxonomy to reduce semantic overlap between closely related error categories. For instance fine-grained categories such as  misunderstanding, neglect, and inexpert statement are very close and can be assigned one single error type. The coarse-grained taxonomy mitigates boundary ambiguity while preserving the core error type signal.
The mapping between original error type definitions in \cite{du2024llms} and our coarse-grained error types and their definition is provided in the appendix document available in the git repository. This step is skipped for non-deficient segments.

\textbf{Rebuttal Action Prediction (RAP):}
Each error type is mapped to one or more possible rebuttal actions. We use the same set of rebuttal actions defined in \citet{Purkayastha2023ExploringJA}. The mapping is provided in \ref{tab:error_type_rebuttal_action_mapping} in the appendix. For the review segment under consideration with an identified deficiency status and an error type (if applicable), LLM  predicts the rebuttal action from the subset of rebuttal actions fetched from the static table. % (Prompt in Table \ref{tab:prompts_dp_etp_rap} in appendix). 
\textbf{Rebuttal generation:}
Finally, the rebuttal segment is generated by prompting the LLM which takes deficiency status, error-type, rebuttal action as input together with the title of the paper, content of the paper, and the review segment.% (Prompt in Table \ref{tab:prompts_sa} in appendix).

\textbf{Utilization of RAG:}
\textit{Retrieval from the paper}: We examine the performance of rebuttal generation by retrieving relevant content from the paper for a particular review segment with an additional LLM call, rather than providing the entire paper. 
\textit{Retrieval from the literature}: We also examine the performance of rebuttal generation by providing relevant literature to respond to certain review segments. We retrieve relevant literature using scholarqa-api \cite{singh2025ai2} which uses semantic scholar\footnote{\url{https://www.semanticscholar.org/product/api}}. We convert the review segment  to a valid query to retrieve the literature relevant to answer the review segment, clarify-doubts, refute, etc.

All the prompts used are provided in the git repository\footnote{\label{fn:defend}\url{https://github.com/jyotsanakhatri24/DefendRebuttalGenerator}}.

\section{\sysname\ : Author Guided Rebuttal Generation}

We  develop  \sysname\ %({Figure \ref{fig:mainpage},\ref{fig:segments},\ref{fig:refinement},\ref{fig:consolidation})
, a web based tool that assist  authors in generating rebuttals for peer-reviews. The system takes as input the paper title, paper content, and a specific review from a reviewer. It segments the review and  generates a rebuttal for each  segment using \textbf{SWRG}. If the author does not agree with the generated rebuttal for any review segment, %it asks a series of questions based on different steps of the pipeline. If the author disagrees, 
the system follows the \textbf{SA} pipeline. It predicts the review deficiency and shows the prediction to the author in a natural language (NL) as illustrated in Table \ref{tab:deficiency_questions}. Demonstrating the output in NL as opposed to providing the class label (deficient or in-deficient) to the author, allows to have more interpretability. A raw class label, lacks the contextual richness necessary for effective author-in-the-loop interaction. The system moves to the next step as either error-type prediction if the review segment is agreed to be deficient or rebuttal action prediction if it agreed to be not deficient. The system predicts the error type and presents the prediction to the author in NL form. The static look-up table of the mapping between the error type classes and the NL responses is provided in Table \ref{tab:errortypestatement}. If the author does not agree to the predicted error type, there is a choice to rectify the same by providing feedback in terms of NL and move to the rebuttal action prediction step. Based on the predicted rebuttal action an appropriate message is shown to the author which can be rectified, if need be (Mappings between the rebuttal actions and the corresponding NL messages are illustrated in Table \ref{tab:rebuttal_action_statement}). Finally, the system  proceeds to final rebuttal generation in context of the finalized deficiency labels, error types and rebuttal actions. The author has the choice to edit the generated rebuttal. At the end, the system  provides the final rebuttal consolidated using all rebuttal segments. 

The tool allows staged approach by decomposing the task into sub-tasks emulating the reasoning steps involved in the rebuttal generation guided by \sysname\, making the task simpler for the author. Moreover, it allows the author to make a selection at every step, which carries less cognitive load as opposed to explicitly mentioning the actions  or directly generating the complete or segment-wise rebuttal from scratch, increasing the efficiency of the author. 
The link of the  video\footnote{\url{https://drive.google.com/file/d/1uTP9j2rUEBdNip0vQ0Z9J_rzsHUA12lr/view?usp=sharing}}, and the code for running the web interface\footref{fn:defend} is provided.

\section{Experiments and Results}

\subsection{Dataset}
We use a subset of the ReviewCritique\cite{du2024llms} dataset for benchmarking various steps of \textbf{SA}, and final rebuttal generation. Considering the high cognitive load and complexity of creation of the fine-grained expert annotated data, we select a subset of 4 papers: 2 accepted and 2 rejected, with total 13 reviews and 185 review segments. While the number of papers is limited, the dataset contains a substantial number of annotated segments. The original dataset has annotations of deficiency status and error types at review sentence level.  We change the segmentation to meaningful semantic units and perform annotations for deficiency, error-type, and rebuttal actions after creating the mapping from each review segment to corresponding rebuttal segments, by following these steps: (i) 
For each review, we retrieve rebuttals using the \url{openreview.net} API, (ii) We create semantic segments of the reviews using an LLM, (iii) For each review segment, a mapping to one or more rebuttal segments is created manually. (iv) For each review segment, to annotate the deficiency and error-type labels were manually annotated. (v) For each rebuttal segment, the rebuttal action is predicted using an LLM providing review segment, and rebuttal segment both as an input together with the paper title and paper content. Then these predictions are provided to an annotator.

There were two annotators, PhD level NLP researchers. One annotator annotated, and other verified the annotations and in case of no consensus they sat together and analyze to reach consensus.
We report pre-adjudication agreement between the annotator and verifier, which was 95\% for review-to-rebuttal mapping, 86\% for deficiency prediction (DP), 74\% for error-type prediction (ETP), and 80\% for rebuttal action prediction (RAP). For rebuttal actions the first annotator started the annotation by looking at the LLM predicted outputs. For review-to-rebuttal mapping, the majority of disagreements stemmed from missed mappings due to granularity mismatches, where a single rebuttal segment responded to multiple review comments. For deficiency prediction, differences emerged in determining whether a concern crosses the threshold from minor issue to major deficiency. For error-type prediction (ETP), most mismatches were due to category overlap. Similarly for the rebuttal action prediction, one rebuttal segment often performs multiple functions simultaneously, disagreements arose when one annotator selected the dominant action. 
\subsection{Results}
%\textcolor{red}{The following table can be single column: not sure what it means}
\begin{table}[]
\centering
\resizebox{\columnwidth}{!}{
\begin{tabular}{ p{2cm} p{2cm} p{1.5cm} p{1.5cm} p{1.5cm}}
    \hline
    \textbf{Metric} & \textbf{Approach} & \textbf{No-RAG (Full Paper)} & \textbf{Paper-Context only)} & \textbf{Lit-Aug} \\
    \toprule
    
    \multirow{3}{=}{Factual Correctness} 
    & DRG & .45 & -- & .42 \\
    & SWRG & .58 & .54 & .58 \\
    & SA-Pred & .62 & .59 & .63 \\
    & \sysname\ & .74 & .71 & .77 \\
    \midrule
    \multirow{3}{=}{Consistency}
    & DRG & .81 & -- & .83 \\
    & SWRG & .90 & .86 & .87 \\
    & SA-Pred & .88 & .86 & .91 \\
    & \sysname\ & .94 & .90 & .90 \\
    \midrule
    \multirow{3}{=}{Strength of Refutation}
    & DRG & .12 & -- & .23 \\
    & SWRG & .29 & .29 & .35 \\
    & SA-Pred & .29 & .29 & .29 \\
    & \sysname\ & .76 & .71 & .76 \\
    \bottomrule
    \end{tabular}
    }
    \caption{Results on Extended ReviewCritique Dataset. Lit-Aug: Literature Augmentation: Segment-level precision for Strength of Refutation, Factual Correctness, and Consistency. SA-Pred: SA pipeline with predicted deficiency, error-type, and rebuttal action, \sysname\ : SA pipeline with gold deficiency, error-type, and rebuttal action.}
    \label{tab:allpipelines}
\end{table}

We use gemini-2.0-flash \cite{gemini2_blog2024} wherever an LLM call is required. We provide evaluation at segment level as opposed to a score for the complete rebuttal. We evaluate the approaches using three aspects strength of refutation, factual correctness, and consistency using LLM-as-a-judge. % Prompts in Table \ref{tab:prompts_consistency}, \ref{tab:prompts_refutation}, \ref{tab:prompts_fc} in appendix.
Strength of refutation represents the ability of the system to refute the criticism whenever needed. Factual correctness represents the ability  to generate factually correct responses without any hallucination or wrong information. Consistency represents the ability  to generate coherent, consistent, non-ambiguous responses without having any contradiction amongst it. %\textcolor{red}{It would be better if we provide the definitions of these three aspects.:Done}
For strength of refutation, we calculate the matching between the strength of refutation of gold rebuttal segments and our approaches and report precision to measure whether the model is able to refute wherever needed. For consistency, and factual correctness we report the precision considering all gold rebuttal segments are factually correct and consistent. Factual correctness is determined by providing gold rebuttal segment in the context. For \textbf{SA}, we use the predicted deficiency status, error-type, and rebuttal action, and for \sysname\ we use the gold deficiency status, error-type, and rebuttal action to generate the rebuttal to examine the pipeline, as in our system the author can choose the right option. For the evaluation of \textbf{DRG} at segment level, we segment the generated rebuttal and then for each review segment identify a rebuttal segment which is applicable for that particular segment. We further follow the same process of segment-wise evaluation. \\

\begin{center}
    \begin{table}[]
    \centering
    \resizebox{\columnwidth}{!}{
    \begin{tabular}{c|c}
    \toprule
    \textbf{Stage} & \textbf{Precision/Recall/F1-score} \\ \midrule
    \textbf{Deficiency Prediction} &  .54/.53/.53 \\ \midrule
    \textbf{Error Type Prediction} & .50/.62/.46 \\ \midrule
    \textbf{Rebuttal action Prediction} & .28/.20/.16  \\ \bottomrule
    \end{tabular}
    }
    \caption{Performance of zero-shot LLMs for various steps of the pipeline in Sequential Approach (\textbf{SA})}
    \label{tab:benchmark_each_step}
    \end{table}
\end{center}

\begin{table*}[t]
\centering
\small
\begin{tabularx}{\textwidth}{lXc}
\toprule
\textbf{Category} & \textbf{Question} & \textbf{Avg. ($\mu \pm \sigma$)} \\
\midrule

\multirow{4}{*}{Rebuttal Quality}
& Q1. The final rebuttal adequately addresses the reviewer’s concerns. & 3.83 $\pm$ .68 \\
& Q2. The rebuttal is factually accurate and well-grounded in the paper. &  3.92 $\pm$ .58\\
& Q3. The rebuttal appropriately refutes incorrect or misleading reviewer claims. & 3.92 $\pm$ .2 \\
& Q4. The rebuttal is coherent and internally consistent. & 3.83 $\pm$ .41  \\

\midrule
\multirow{3}{*}{Controllability}
& Q5. I could easily understand why the system generated a particular rebuttal. & 3.92 $\pm$ .66 \\
& Q6. The intermediate steps were useful for guiding the rebuttal. & 4.00 $\pm$ .84 \\
& Q7. I felt in control of the final rebuttal content. & 4.08 $\pm$ .92 \\

\midrule
\multirow{3}{*}{Cognitive Load}
& Q8. The interaction with the system was easy to follow. & 4.67 $\pm$ .52 \\
& Q9. The system reduced the effort and time required to write a rebuttal. & 4.25 $\pm$ .88 \\
& Q10. The system reduced my cognitive load compared to writing manually. & 4.25 $\pm$ .88 \\

\midrule
Overall Usefulness
& Q11. I would use this tool for preparing real rebuttals. &  3.80 $\pm$ .45 \\

\bottomrule
\end{tabularx}
\caption{Average ratings across users (5-point Likert scale: 1 = Strongly disagree, 5 = Strongly agree).}
\label{tab:user_study_results}
\end{table*}

\subsection{Discussion}
%Highlight the improvement in refutation and factual correctness
The poor performance of \textbf{DRG}, in Table \ref{tab:allpipelines} shows that rebuttal writing is not a simple generative task. \textbf{SWRG} and \textbf{SA} helps in improving the consistency, factual correctness and strength of refutation. Segmenting reviews into semantically coherent units helps constrain the generation process and allows the system to focus on localized concerns, while the \textbf{SA} introduces explicit reasoning and helps in responding to critiques in review segments. This shows the importance of structure, decomposition, and explicit reasoning in rebuttal generation. In our experiments when we use retrieval from the paper, we do not observe improvements. Whereas when we augment relevant literature, we observe slight improvement for \textbf{SA} in terms of factual correctness.

We also benchmark each step of the pipeline (DP, ETP, and RAP) using precision in Table \ref{tab:benchmark_each_step}.  As it can be observed, LLM finds these tasks very challenging leading to very low performance, hinting at the need of author intervention with minimal cognitive overload. 

Finally, our demo shows this into practice, by first providing \textbf{SWRG} based drafts and then allowing  the author to guide the tool through \textbf{SA} pipeline by providing inputs only when necessary. This allows the author to steer the rebuttal generation with minimal intervention,  minimizing his cognitive burden and thus improving the efficiency along with the accuracy .

\subsection{User Study}
To assess the practical utility of \sysname\, we conduct a user study evaluating rebuttal quality (in-terms of strength of refutation, factual correctness, consistency, and completeness), controllability, cognitive load, and overall helpfulness. We had N = 6 participants (researchers in NLP and ML) with prior experience in submitting papers to major NLP conferences. All participants had written at least one rebuttal in the past. Participants interacted with the Defend web interface and rated their experience using a 5-point Likert scale (1 = Strongly disagree, 5 = Strongly agree). Questionnaire and average rating across users is provided in Table \ref{tab:user_study_results}. 
%Rebuttal quality
Participants rated the system highly on all aspects of rebuttal quality.
%Strength of refutation
High scores on strength of refutation indicate that the system successfully identifies and responds to incorrect or misleading reviewer claims.
%Controllability
It also received positive feedback on controllability (Q5–Q7). Participants reported that they found the intermediate steps (deficiency identification, error type classification, rebuttal action) useful for guiding revisions.
%Cognitive load
Scores on cognitive load (Q8–Q10) indicate that interaction with \sysname\ was easy to follow and reduced both effort and time required to draft rebuttals.
%Overall
The agreement with Q11 validates the system’s practical applicability.

\section{Conclusion}

In this paper, we introduce \sysname\, an LLM based tool designed to explicitly execute the underlying reasoning process of automated rebuttal generation, while keeping the author-in-the-loop. We present a structured mechanism for generating rebuttals using LLMs. We extend a subset of the ReviewCritique dataset to include deficiency status, error-type, rebuttal action labels and gold rebuttal segments. We conduct a comparison of three paradigms of rebuttal generation: \textbf{DRG}, \textbf{SWRG}, and \textbf{SA}. Our experimental findings reveals that \textbf{SA} with minimal author intervention substantially improve factual correctness and strength of refutation. We observe retrieving relevant literature and including it while generating rebuttal helps in improving the factual correctness but not strength of refutation.

\section*{Limitations}

Our evaluation is performed using LLM-as-a-judge.% on a small scale dataset. 
Although, we have provided the gold rebuttal in the context and using LLM as a judge  is a common practice, this may introduce biases, and may not be able fully capture scientific nuances. To retrieve relevant literature, more effective grounding techniques are required. 

\section*{Ethical Considerations}
We use only publicly available peer-review data and do not release any private or confidential reviewer content. Our system is designed strictly as an assistive tool with required human oversight. 

\bibliography{custom}

\appendix

\section{Appendix}
\label{sec:appendix}

\begin{table*}
\centering
\resizebox{\textwidth}{!}{
\begin{tabular}{p{5cm}|p{5cm}|p{7cm}}
\toprule
\textbf{Error-type} & \textbf{Error-type-finegrained} & \textbf{Definition}  
\\ \midrule
Incorrect references & Invalid Reference, Concurrent work & The reviewer is not citing the appropriate sources (non peer-reviewed or concurrent work) in the current statement.
\\ \midrule
Less rigor in reviewing methodlogy and experiments & Out-of-scope, Invalid Criticism, Less rigor in reviewing methodology and experiments
 & The reviewer is suggesting things beyond the scope of the paper or the reviewer's criticism is invalid.
\\ \midrule
Misinterpretation of claims and ideas presented in the paper & Misunderstanding, Neglect, Inexpert Statement, Unstated statement
 & The reviewers is misinterpreting the claims and ideas presented in the paper and overlooked important details of the paper or the reviewer is exhibiting lack of domain knowledge or not supported by the content of the paper.
\\ \midrule
Superficial and vague review & Misinterpret Novelty, Vague Critique,
Subjective, Superficial Review, Missing Reference & In the current statement, the reviewer has misinterpreted novelty or the reviewer is lacking specificity of the components.
\\ \midrule
Incomplete, incorrect, or copied summary & Inaccurate Summary, Summary Too Short, Copy-pasted Summary & The summary is misrepresenting the content of the paper, or too short or directly copied from the paper.
\\ \midrule
Syntactic, structural, or semantic issues in the review & Typo, Contradiction, Misplaced attributes, Duplication & The review segment has typological errors that are affecting the clarity.
\\ \midrule
Misidentification of Syntactic or structural issues in the paper & Writing, Misunderstanding of the Submission Rule & The reviewer has misidentified the structural issues in the paper.
\\ \bottomrule
\end{tabular}
}
\caption{Error type and their definitions}
\label{tab:error_type_definition}
\end{table*}

\begin{table*}[]
    \centering
    \resizebox{\textwidth}{!}{
    \begin{tabular}{p{6cm}|p{10cm}}
    \toprule
    \textbf{Error-Type} & \textbf{Possible set of Rebuttal Actions}
    \\ \midrule
    Incorrect references & Reject Request
    \\ \midrule
    Less rigor in reviewing methodology and experiments & Accept for future work, Reject criticism, Refute question
    \\ \midrule
    Misinterpretation of claims and ideas in the paper & Contradict Assertion, Refute question, Reject criticism
    \\ \midrule
    Superficial and vague review & Refute question, Reject criticism, Contradict assertion
    \\ \midrule
    Incomplete, incorrect or copied summary & All
    \\ \midrule
    Syntactic, structural, and semantic issue in the paper & Reject criticism
    \\ \midrule
    Non-deficient & 
    Answer question, The task is done, The task will be done, Concede criticism, Mitigate criticism, Accept praise%Question (Answer question), Request (The tasks is done or will be done), Negative comment (Concede criticism or mitigate criticism), Praise (Accept praise)
    \\ \bottomrule
    \end{tabular}
    }
    \caption{Static mapping of error-type to rebuttal actions}
    \label{tab:error_type_rebuttal_action_mapping}
\end{table*}

\begin{table*}[h]
\centering
    \begin{tabular}{p{4cm} p{11cm}}
    \hline
    \textbf{Deficiency status} & \textbf{Question} \\
    \hline
    True &
    The review statement is not valid. It contains either factual errors, lacks constructive feedback, is subjective, or is without evidence (Deficient). \\
    Falsse &
    The review statement is valid in terms of factuality and constructive feedback (Non-deficient).\\
    \hline
    \end{tabular}
    \caption{Natural language statement mapping for deficiency status}
    \label{tab:deficiency_questions}
\end{table*}

\begin{table*}[h]
\centering
\begin{tabular}{p{4cm} p{11cm}}
\hline
\textbf{Error-type} & \textbf{Statement} \\
\hline
Incorrect references & The reviewer is not citing the appropriate sources (non peer-reviewed or concurrent work) in the current statement.\\
Less rigor in reviewing methodology and experiments & In the current statement, the reviewer is suggesting things beyond the scope of the paper or the reviewer's criticism is invalid. \\
Misinterpretation of claims and ideas presented in the paper & In the current statement, the reviewer is misinterpreting the claims and ideas presented in the paper and overlooked important details of the paper or the reviewer is exhibiting lack of domain knowledge or not supported by the content of the paper. \\
Superficial and vague review & In the current statement, the reviewer has misinterpreted novelty or the reviewer is lacking specificity of the components. Do you agree? \\
Incomplete, incorrect, or copied summary & In the current statement, the summary is misrepresenting the content of the paper, or too short or directly copied from the paper.\\
Syntactic, structural, or semantic issues in the review & The current review statement has typographical errors that are affecting the clarity. \\
Misidentification of syntactic or structural issues in the paper & In the current review statement, the reviewer has misidentified the structural issues in the paper. \\
\hline
\end{tabular}
\caption{Natural language statement mapping for error types}
\label{tab:errortypestatement}
\end{table*}

\begin{table*}[h]
\centering
\begin{tabular}{p{4cm}|p{11cm}}
\hline
\textbf{Rebuttal action} & \textbf{Statement} \\
\hline
    Reject work & The request needs to be rejected. \\
    Accept for future work & The suggested things needs to be accepted as future work. \\
    Reject criticism & The criticism needs to be rejected. \\
    Refute question & The question needs to be disproved. \\
    Contradict assertion & The statement needs to be contradicted. \\
    Mitigate criticism & The rebuttal statement needs to be generated in a manner that represents the statement is not important. \\
    Answer question & The question needs to be answered. \\
    The task is done & The rebuttal statement needs to specify the task has already been done and pinpoint where. \\
    The task will be done & The rebuttal statement needs to specify the task will be done in camera ready. \\
    Concede criticism & The rebuttal statement needs to admit to the provided criticism. \\
    Accept praise & The rebuttal statement needs to accept the praise. \\
\hline
\end{tabular}
\caption{Natural language statement mapping for rebuttal actions}
\label{tab:rebuttal_action_statement}
\end{table*}
%\subsection{Appendices}

%\section{Bib\TeX{} Files}
%\label{sec:bibtex}

%Global structure of the rebuttal

\end{document}